\documentclass[11pt]{wlscirep}
\usepackage[utf8]{inputenc}
\usepackage{hyperref}
\usepackage{bm}
\usepackage{epstopdf}
\usepackage{makecell}
\usepackage{graphicx}
\usepackage{lscape}
\usepackage{pdflscape}
\usepackage{multirow}
\usepackage{amsmath}
\usepackage{makecell}
\usepackage{times}
\usepackage{epsfig}
\usepackage{amssymb}
\usepackage{threeparttable}  
\usepackage{float}
\usepackage{epsfig}
\usepackage{balance}
\usepackage{booktabs}
\usepackage{enumitem}
\usepackage{mathtools}
\usepackage{makecell}
\usepackage{multirow}
\usepackage{pifont}
\usepackage{stackengine}
\usepackage{xcolor}
\usepackage{colortbl}
\usepackage{tabu}
\usepackage{geometry}
\usepackage{fancyhdr}
\usepackage{array}
\usepackage{subcaption}
\usepackage{graphicx}
\usepackage{caption}
\usepackage{longtable}
\usepackage{color}
\usepackage{cleveref}   
\usepackage{xspace}
\usepackage[breakable]{tcolorbox}
\usepackage{titletoc}
\usepackage{setspace} 
\usepackage{amsfonts}  

\title{\fontsize{18pt}{16pt}\selectfont From Occurrence to Consequence: A Comprehensive Data-driven Analysis of Building Fire Risk}

\newcommand{\be}{\begin{equation}}
\newcommand{\ee}{\end{equation}}
\newcommand{\ba}{\begin{align}}
\newcommand{\ea}{\end{align}}

\def\ba{{$\bm{a}$}}

\definecolor{myblue}{RGB}{0, 133, 186}
\definecolor{mygreen}{RGB}{72, 194, 0}
\definecolor{myred}{RGB}{177, 0, 28}
\definecolor{good}{rgb}{0.11, 0.77, 0.11}
\definecolor{bad}{rgb}{0.77, 0.11, 0.11}

\author[1+]{Chenzhi Ma}
\author[1,2+]{Hongru Du}
\author[3]{Shengzhi Luan}
\author[4]{Ensheng Dong}
\author[1,2,5]{Lauren M. Gardner}
\author[1*]{Thomas Gernay}

\affil[1]{Department of Civil and Systems Engineering, Johns Hopkins University, Baltimore, MD, USA.}
\affil[2]{Center for Systems Science and Engineering, Johns Hopkins University,
Baltimore, MD, USA.}
\affil[3]{Department of Mechanical Engineering, Boston University, Boston, MA, USA.}
\affil[4]{RAND Corporation, Arlington, VA, USA.}
\affil[5]{Department of Epidemiology, Johns Hopkins Bloomberg School of Public Health, Baltimore, MD, USA}

\small{\affil[+]{These authors contributed equally.}
\affil[*]{The corresponding author information: tgernay@jhu.edu}

\begin{abstract}
\fontsize{11pt}{12pt}\selectfont

Building fires pose a persistent threat to life, property, and infrastructure, emphasizing the need for advanced risk mitigation strategies. This study presents a data-driven framework analyzing U.S. fire risks by integrating over one million fire incident reports with diverse fire-relevant datasets, including social determinants, building inventories, weather conditions, and incident-specific factors. By adapting machine learning models, we identify key risk factors influencing fire occurrence and consequences.
Our findings show that vulnerable communities, characterized by socioeconomic disparities or the prevalence of outdated or vacant buildings, face higher fire risks. Incident-specific factors, such as fire origins and safety features, strongly influence fire consequences. Buildings equipped with fire detectors and automatic extinguishing systems experience significantly lower fire spread and injury risks.
By pinpointing high-risk areas and populations, this research supports targeted interventions, including mandating fire safety systems and providing subsidies for disadvantaged communities. These measures can enhance fire prevention, protect vulnerable groups, and promote safer, more equitable communities.

\vspace{1em}

\textbf{keywords}: community risk analysis, building fire risk, data-driven, fire occurrence, fire consequences
\end{abstract}

\begin{document}
\flushbottom
\maketitle
\thispagestyle{empty}
\def\graphmodality{detector graph modality}
\def\gridmodality{GPS grid modality}

\newpage

\section{Introduction} 
\label{sec:intro}

Fire has long shaped human civilization, enabling development while posing a persistent threat. From ancient fire breaks to modern suppression systems, humanity has sought to control this dual-edged force, which has enabled development and survival but also brought catastrophic losses when uncontrolled\cite{yu2020quantifying, hantson2022human, solomon2023chlorine,zhao2024forest}. Yet, building fires remain among the most devastating, directly impacting lives and properties in the spaces where people live and work \cite{franchini2024integrating, bowman2009fire, tang2024epidemiology, bonner2024examining}. In the U.S. alone, building fires accounted for 77\% of civilian fire deaths, 88\% of civilian fire injuries, and 83\% of direct property damage in 2022 \cite{hall2022fire}. Despite substantial annual financial investments exceeding \$57 billion in fire safety measures \cite{zhuang2017total}, such as sprinklers, detectors, and structural fireproofing, the significant reduction in fire incidents and their associated consequences throughout the 20th century has plateaued over the last two decades \cite{hall2022fire}, with no notable change observed during this period. This stagnation raises a critical question: \textit{Why have modern fire safety efforts failed to further mitigate fire risks?} 

Fire risk encompasses both the probability of a fire occurrence and its potential consequences. Modern fire risk analysis frameworks combine historical data with probabilistic methods to estimate the likelihood of fire ignition \cite{van2019need, meacham2004understanding, hu2016research, macleod2020reliability}, which is often assumed to depend on factors such as building occupancy and total floor area. Once a fire is ignited, numerical and experimental models are employed to simulate fire growth \cite{hu2024experimental,dundar2023fire,mozaffari2023real}, material or structural responses under extreme heat \cite{ma2024numerical, muniraj2024mechanical,zhao2023review,choe2021fire,ni2021framework}, and evacuation behaviors and expected casualties \cite{ding2024building,fu2024framework,zhang2023research,xiao2022evacu,haghani2024human}. Current frameworks are instrumental in assessing fire consequences and informing fire safety designs for buildings, while they often treat fires as isolated events, focusing narrowly on individual buildings or predefined scenarios. 
Recently, a broader focus on community fire risk analysis has gained attention \cite{NFPA1300}. Regional and local studies have revealed that fire risks vary significantly across communities, with inequalities linked to community characteristics \cite{turner2017risk,schmidt4757205machine,hossain2023analyzing,xia2019detecting}. Simultaneously, previously overlooked building-level factors (e.g., ignition sources) are increasingly recognized for their considerable impact on fire outcomes \cite{challands2010relationships, troitzsch2016fires, tali2017location}. 
However, existing research on fire risk analysis often focuses on isolated factors or localized contexts, underestimating the complex interplay between community-level vulnerabilities and incident-specific factors. This limited scope hinders the development of evidence-based risk assessments and broadly applicable mitigation strategies. Addressing this gap requires an integrated approach that accounts for both broader community risks and localized fire event characteristics.




In this study, we conduct a comprehensive, data-driven analysis of fire risk across the U.S., examining the full spectrum of building fire progression from occurrence to consequences. We investigate the spatiotemporal distribution of fire incidents nationwide, identify key factors influencing both fire occurrence and resulting consequences, and discuss potential strategies for effective risk mitigation. Our analysis leverages a novel dataset integrating over one million building fire incidents with detailed fire incident reports, social determinants, business compositions, building inventories, and weather conditions, providing coverage for over 90\% U.S. counties and all 50 states.
Our analysis reveals a strong association between localized vulnerabilities and elevated fire risk: communities with higher vulnerability in social determinants and building inventories face higher fire occurrence, fire spread, and injury risks. Moreover, cooler and dryer weather conditions tend to elevate fire risk in both occurrence and consequences. These findings highlight the importance of making fire prevention efforts with the consideration of unique local factors, such as organizing fire safety education campaigns in under-educated communities.
Critically, our analysis reveals that fire-related factors (such as ignition source and structural characteristics) significantly affect the fire spread and economic loss risks, while human-related factors (such as demographic composition) significantly affect the injury risks. This observation supports the development of target-specific strategies, such as developing advanced evacuation strategies and mandating Automatic Extinguishing Systems (AES) for buildings with a high ratio of vulnerable populations, enabling more effective and equitable investment in fire prevention. 


\section{Results}
\label{sec:results}

\begin{figure}[H]
    \centering
    \includegraphics[width=0.9\linewidth]{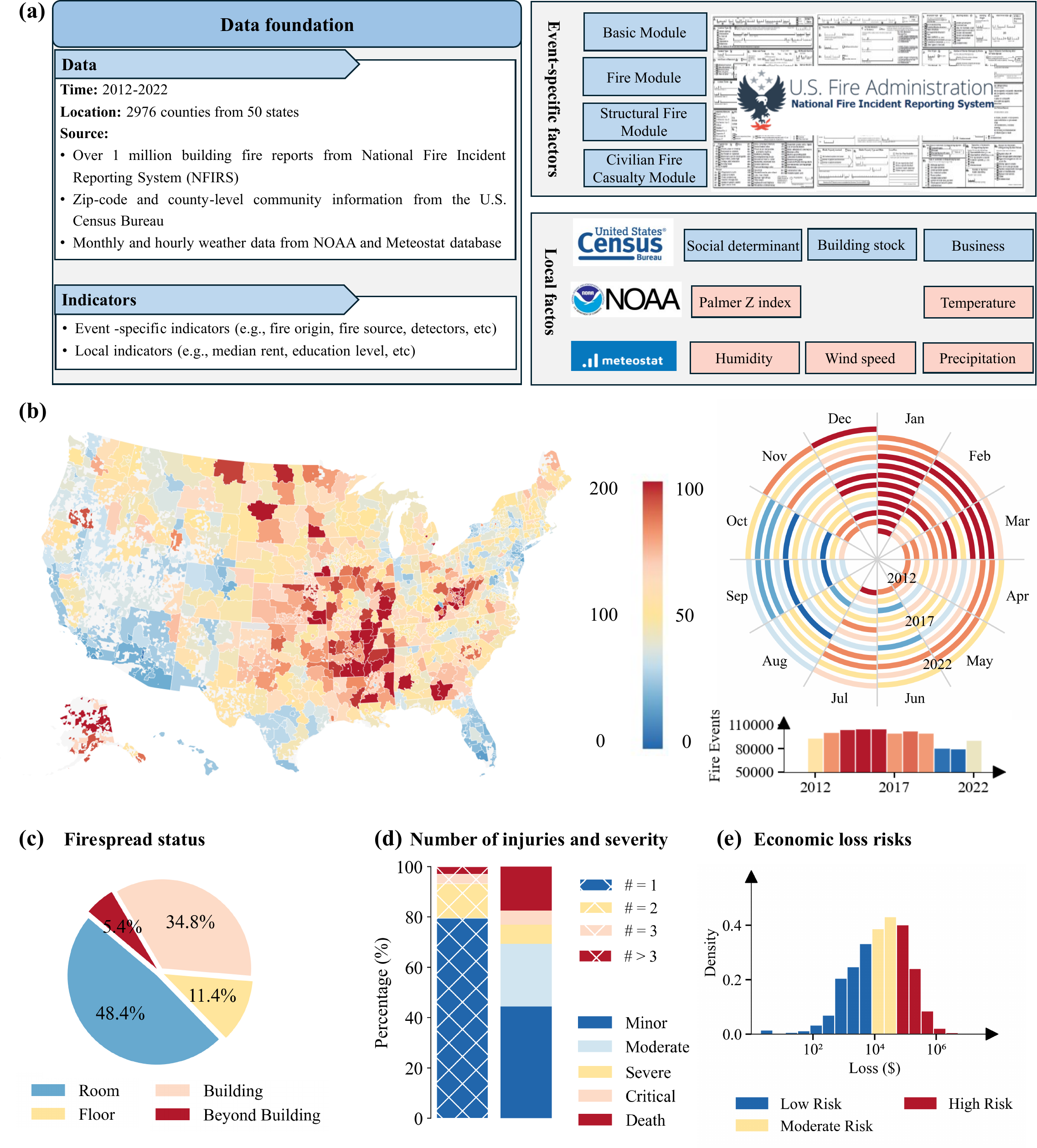}
    \caption{\small \textbf{Data foundation for nationwide building fire risk analysis.} \textbf{a} The building fire-related dataset combines over one million NFIRS fire incident reports (2012–2022) with detailed incident-specific information on building characteristics, incident timing and location, firefighting efforts, and local information on demographics, social determinants, building inventories, business proportions, and weather conditions at monthly and hourly resolution. \textbf{b} Spatial distribution of building fire incident rates, defined as the number of fires spreading beyond the item of origin per 100,000 building units annually. The spiral chart illustrates the monthly distribution of building fire events across the U.S., with data normalized relative to the peak fire month of each year.  \textbf{c} Classification of fire spread from the room of origin: confined to the room, floor, building, or beyond the building of origin.  \textbf{d} Distribution of fire-induced injuries for incidents with reported injuries, including number of injuries and severity level. \textbf{e} Distribution of fire-induced economic loss, with classification into low, moderate, and high based on quantile thresholds at 40\%, 75\%, and 100\% of the historical loss data.}
    \label{fig:overall}
\end{figure}

\subsection{Data foundations for understanding fire risks}

This study constructs a comprehensive building fire-related dataset by integrating over one million building fire incidents, spanning over 90\% of U.S. counties between 2012 and 2022, from multiple independent sources. This nationwide compilation (Fig. \ref{fig:overall}a) uniquely synthesizes five categories of information: (1) detailed fire incident reports, (2) social determinants, (3) business proportions, (4) building inventories, and (5) weather conditions. 

The fire incident reports, drawn from the National Fire Incident Reporting System (NFIRS) \cite{NFIRScitation}, provide rich, incident-level building characteristics (e.g., property usage, detector status), fire attributes (e.g., ignition source, location of fire origin), and emergency response actions (e.g., response times, firefighting tactics). Social determinants, encompassing demographics and socioeconomic variables, along with business proportions and building inventories, are sourced from the US Census Bureau\cite{USCensusBureau} at ZIP-code level. Lastly, we integrated weather conditions from National Oceanic and Atmospheric Administration (NOAA) \cite{noaa_climate_data} and Meteostat \cite{meteostat}, capturing monthly and hourly weather information such as temperature and moisture at the time and location of each fire event. Notably, all the ZIP code level data can also be aggregated to higher spatial resolutions, such as the county level, allowing for broader regional analyses. Detailed descriptions of these variables are provided in the Method Section \ref{data} and Supplementary Section 1.


\subsection{The spatiotemporal landscape of U.S. building fires}
\label{sec:overview}



Fire incident rates, measured as the number of incidents per 100,000 building units (Fig. \ref{fig:overall}b), highlight a disproportionate burden across regions, with significantly higher fire frequencies in parts of the midwest and southeast. The annual number of fire events also fluctuates over time, showing a distinct dip in 2020 and 2021 (Fig. \ref{fig:overall}b). Additionally, the monthly distribution of fire incident rates reveals a clear seasonal pattern, with significantly higher frequencies during winter (Fig. \ref{fig:overall}b). These spatial and temporal variations in fire occurrence underscore the need to explore how local factors, including weather conditions, influence fire occurrence across diverse regions and seasons.

Beyond fire occurrence, the consequences of building fires are assessed from three critical angles (Figs. \ref{fig:overall}c, d, e): fire spread status, human injury, and economic loss. fire spread status is represented by four levels \cite{NFIRScitation}: 1) confined to the room of origin, 2) confined to the floor of origin, 3) confined to the building of origin, and 4) extending beyond the building of origin. 
Human injury is assessed by considering both the number of fire-induced injuries and the severity of those injuries. Lastly, the economic loss is measured through firefighters' estimates of fire-induced financial damage, averaging \$70,000 per incident in 2022 value. 

\subsection{Drivers and disparities in U.S. building fire occurrance}

We utilize three sets of generalized additive models (GAMs) to quantify the relationships between fire incident rates and associated factors at the county level across the U.S. Factors include social determinants, business proportions, building inventories, and weather conditions (see Supplementary Section 1 and Section 2 for complete variables and methods). First, a single national GAM (model diagnostic shown in Supplementary Fig. S4) is fitted to the entire national dataset to establish a broad baseline. Next, four seasonal GAMs (spring, summer, autumn, and winter) are used to capture seasonal variations in incident rates. Finally, four regional GAMs (see Supplementary Fig. S2 and S3 for the boundary of Northeast, Midwest, South, and West) are applied to identify geographical-based risk factors.

\begin{figure}[htpb]
    \centering
    \includegraphics[width=0.9\linewidth]{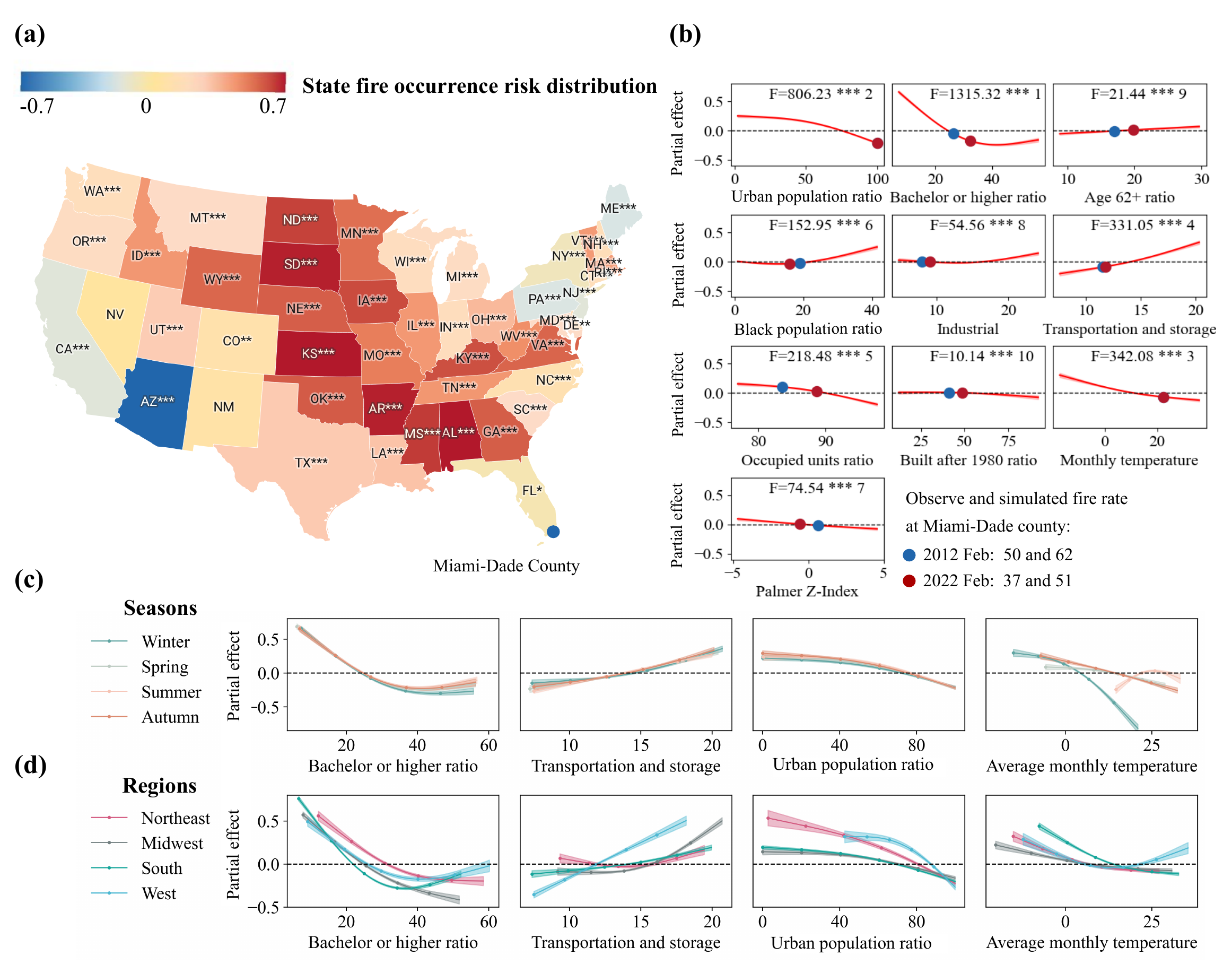}
    \caption{\small \textbf{Fire incident rates in the U.S. and corresponding effects of local factors.} \textbf{a} Fixed effects of states on the fire occurrence in the GAM model, where a darker shade of red signifies a higher fire incident rates, and a darker shade of blue denotes a lower rates. \textbf{b} Partial dependence plots of the local (socioeconomic, building inventory, and business) and climate factors influencing building fire incident rates (n = 38804, excluding counties with events less than 3 in a month). ‘***’: variable significant at p < 0.001. ‘**’: variable significant at p < 0.01. ‘*’: variable significant at p < 0.05. The number on the right indicates the rank of the F-value. 
    Miami-Dade County is highlighted on the plot with a blue dot. Results from seasonal (sub-figure \textbf{c}) and regional (sub-figure \textbf{d}) GAMs incorporating the most significant variables identified within each category: social determinant, business proportion, building inventory, and weather conditions. Details of the GAMs are provided in the Methods Section \ref{sec:occurrences}, with the complete results included in the Supplementary Section 3.}
    \label{fig:step1_occ}
\end{figure}

The building fire incident rates vary both temporally and spatially. As shown in Fig. \ref{fig:overall}b, fire incident rates fluctuate seasonally and annually, with colder months consistently experiencing higher fire incident rates. Additionally, there is notable spatial variation: Fig. \ref{fig:step1_occ}a illustrates state-level effects from the national GAM, using Alabama as the baseline (state effect set to 0). Elevated fire incident rates are evident in the Midwest and Southeast regions, such as Kansas and Kentucky, while states in the Southwest and Pacific regions, including California and Arizona, exhibit lower fire incident rates. These temporal and spatial patterns motivate us to extend our analysis to seasonal and regional GAMs to better understand how seasonal dynamics and regional heterogeneity shape fire incident rates.




Our analysis reveals that vulnerable communities face higher fire incident rates.
Fig. \ref{fig:step1_occ}b highlights strong associations between fire incident rates and indicators of community vulnerability:
1) Under-educated communities: counties with fewer residents holding a bachelor's degree or higher are at greater fire incident rates. This association remains consistent across both seasonal and regional GAM analyses (Figs. \ref{fig:step1_occ}c, d), underscoring the persistent impact of educational disparities on fire vulnerability. 
2) Senior communities: counties with a higher proportion of residents aged 62 and older also face elevated fire incident rates, though this trend is weaker in the South (see Supplementary Fig. S3). 
3) Minority communities: counties with Black populations exceeding 15\% exhibit higher fire rates, particularly in the southern region. Conversely, counties with Black populations below 10\% tend to show reduced fire incident rates (see Supplementary Figs. S3, S4).


Counties with a large proportion of commercial buildings face higher fire incident rates.
The elevated fire occurrence is driven by specific types of economic activity:
1) Logistics hubs: counties with a higher proportion of transportation and storage establishments face increased fire incident rates, likely linked to the storage and handling of combustible materials. These counties often serve as regional distribution and supply chain hubs, where the storage of combustible materials and extensive use of heavy equipment heighten fire hazards. This association remains consistent across seasons and regions; however, in the Northeast, the relationship is relatively flat compared to other areas (Fig. \ref{fig:step1_occ}c). 
2) Manufacturing hubs: counties with a significant manufacturing presence (higher proportion of industrial establishments) also show elevated fire rates. While this trend is consistent across seasons, it differs in the Midwest region, where a higher proportion of manufacturing establishments is associated with lower fire rates (see Supplementary Fig. S3). 
3) Rural areas and sparsely occupied neighborhoods face higher fire incident rates. The negative association between the "Urban population ratio" and building fire incident rates highlights the heightened vulnerability of rural areas (Fig. \ref{fig:step1_occ}b). This pattern remains consistent across seasons and regions. Likewise, neighborhoods with fewer occupied units face relatively higher fire incident rates. While this trend is observed broadly, it appears less pronounced in the Northeast, potentially reflecting regional differences in housing characteristics or fire prevention measures (Fig. \ref{fig:step1_occ}c).

Aridity and cold temperatures elevate fire incident rates. As shown in Fig. \ref{fig:step1_occ}b, counties with lower monthly average temperatures exhibit higher fire incident rates, likely due to increased reliance on heating sources during colder months. This trend, consistent across all regions, peaks in winter and diminishes in summer (Fig. \ref{fig:step1_occ}c), highlighting seasonal variation in fire frequencies. Furthermore, drier-than-normal conditions, as indicated by lower monthly Palmer Z-Index values, also contribute to elevated fire rates (see Supplementary Figs. S2, S3). This increased vulnerability stems from the higher likelihood of fire ignition and spread under such dry conditions, particularly in areas with abundant vegetation or flammable building materials. The combined impact of these factors, aridity and cool temperatures, creates a heightened fire frequency, especially in counties where both conditions prevail. 


\subsection{Performance of the FireCat in predicting fire consequences}
\label{firecat_performance}
Beyond fire occurrence, we develop FireCat, a tree-based model to predict three types of consequences: fire spread, human injury, and economic loss. Fire spread status is assessed using the predefined categories as listed in section \ref{sec:overview}, while human injury and economic loss risk levels are categorized into high, medium, and low risk. Specifically, human injury is determined by combining the number and severity of civilian injuries. Economic loss is categorized based on thresholds in fire loss amount. Detailed descriptions of how each target variable is derived and processed are available in Method Section \ref{method:step2_targets} and Supplementary Section 2.

We evaluate the model’s performance using seven metrics: accuracy, F-1 score, precision, mean squared error (MSE), weighted mean squared error (WMSE), Brier score, and rank probability score (RPS)\cite{du2024advancing}. The first four metrics quantify point prediction accuracy, while the last two assess the quality of the model’s probabilistic predictions. As a baseline for comparison, we implement a probabilistic model based on the target variable's distribution in the training data. Details about the probabilistic error metric and definition of the baseline can be found in Method Section \ref{step2_study_design}. 

\begin{table}[H]
    \small
    \begin{center}
    \caption{\small \textbf{A summary of models' performances for FireCat and Baseline Model by different targets.} $\uparrow$/$\downarrow$ indicates that higher/lower metric values represent better performance. The bolded values correspond to the model with better performance.}
    \label{tab:performance}
    \setlength{\tabcolsep}{4pt} 
    \renewcommand{\arraystretch}{1.1} 
    \resizebox{0.9\textwidth}{!}{
    \begin{tabular}{c|c|c|c|c|c|c|c|c}
    \toprule[1.5pt]
     \multirow{2}{*}{\makecell[c]{\textbf{Target}}} & \multirow{2}{*}{\textbf{Model}} & \multicolumn{6}{c}{\textbf{Evaluation Metric}} \\
     \cline{3-9}
      & & \textbf{Accuracy $\uparrow$} & \textbf{F1-score$\uparrow$} & \textbf{Precision$\uparrow$} & \textbf{MSE $\downarrow$} & \textbf{WMSE $\downarrow$}  & \textbf{Brier Score $\downarrow$} & \textbf{RPS$\downarrow$}
      \\
      \hline
      
\multirow{2}{*}{Fire spread} & Baseline & 0.502 & 0.335 & 0.252 & 1.893 & 1.804 & 0.718 & 0.194 \\
& FireCat & \textbf{0.573} & \textbf{0.581} & \textbf{0.606} & \textbf{1.005} & \textbf{1.429}  & \textbf{0.545} & \textbf{0.135}\\
\cline{2-9}
\hline
\multirow{2}{*}{Fire injury} & Baseline & 0.406 & 0.234 & 0.165 & 1.272 & 1.202 & 0.649 & 0.208\\
& FireCat & \textbf{0.552} & \textbf{0.538} & \textbf{0.549} & \textbf{0.750} & \textbf{0.909} & \textbf{0.549} & \textbf{0.168}\\

\hline
\multirow{2}{*}{Fire loss} & Baseline & 0.400 & 0.228 & 0.160 & 1.351 & 1.257 & 0.655 & 0.214\\
& FireCat & \textbf{0.564} & \textbf{0.559} & \textbf{0.559} & \textbf{0.677} & \textbf{0.904} & \textbf{0.538} & \textbf{0.164}\\
\cline{2-9}

\cline{1-9}
\bottomrule[1.5pt]
    \end{tabular}
    }
    \end{center}
\end{table}

\begin{figure}[htpb]
    \centering
    \includegraphics[width=0.95\linewidth]{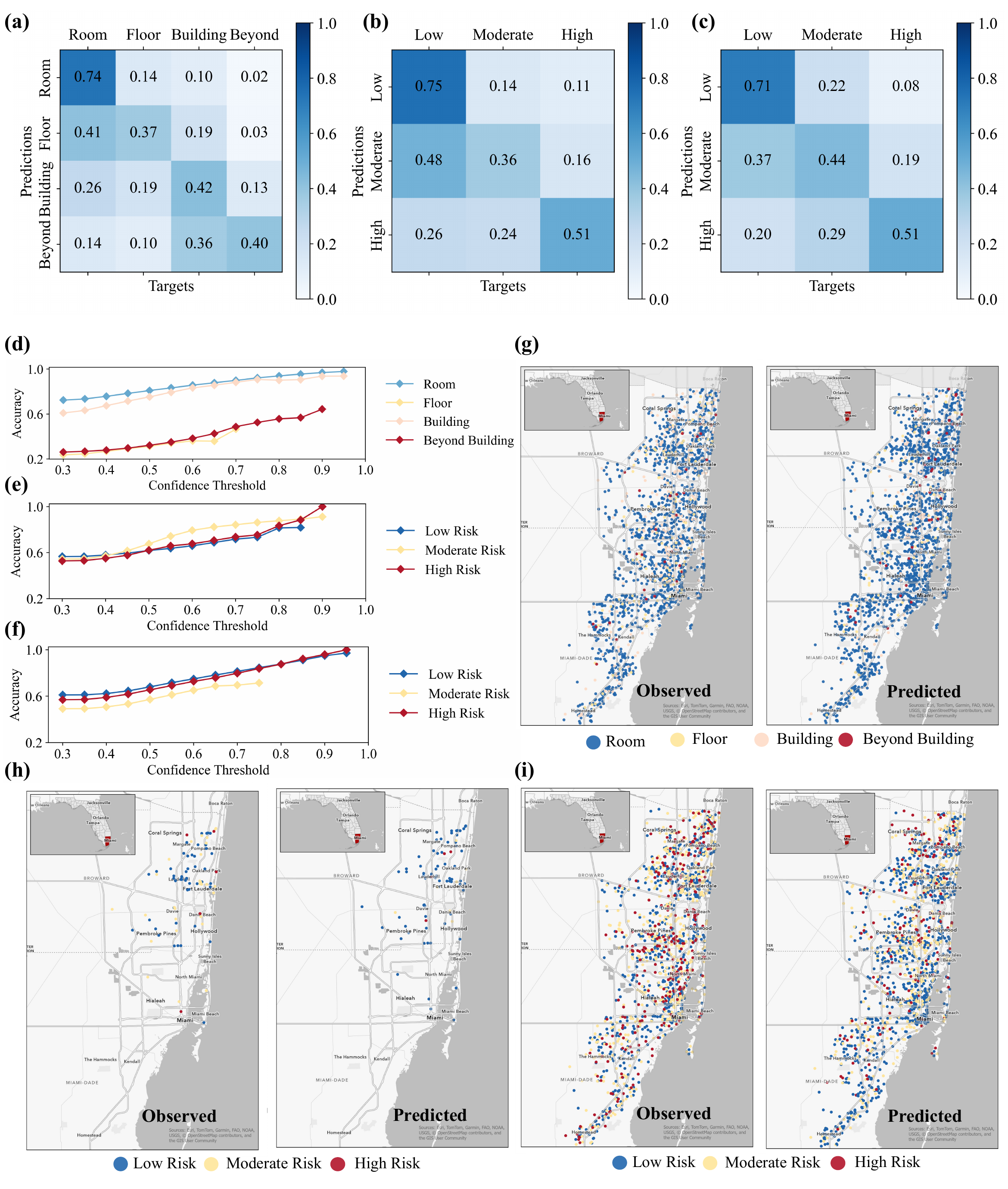}
    \caption{\small \textbf{Performance evaluation and predictions comparison of FireCat across fire spread, human injury, and economic loss.} Confusion matrix and prediction accuracy across various levels of prediction confidence of fire consequence models regarding fire spread (sub-figure \textbf{a, d}), human injury (sub-figure \textbf{b, e}), and economic loss (sub-figure \textbf{c, f}). \textbf{(g, h, i)} Comparison between the predicted and true distribution of fire consequences in Miami-Dade County and Broward County regarding fire spread (sub-figure \textbf{g}), human injury (sub-figure \textbf{h}), and economic loss (sub-figure \textbf{i}).}
    \label{fig:step2_accuracy}
\end{figure}
 
FireCat generates accurate and robust fire consequence predictions. Table \ref{tab:performance} compares the performance of FireCat with the baseline model, demonstrating that FireCat outperforms the baseline across all error metrics with an average accuracy improvement of 30.37\%.


The confusion matrices demonstrate our model's strong performance in predicting the boundary categories (Figs. \ref{fig:step2_accuracy}a, b, c), such as fires contained within a single room or those spreading to or beyond the whole building, as well as in distinguishing between low and high injury and economic loss situations. Errors primarily occur between adjacent risk categories, such as between "Moderate" and "Low." Furthermore, we demonstrate that the confidence level of predictions, represented by the highest probability assigned to a predicted category, serves as a robust indicator of reliability (Figs. \ref{fig:step2_accuracy}d, e, f). As we increase the confidence threshold, wherein only predictions surpassing this criterion are considered, notable improvements in prediction accuracy are observed for all three targets. For instance, at a confidence threshold of 0.9, the model attains an accuracy of over 80\% for predicting whether the fire remains within the room or spreads to the whole building. These results are invaluable for local fire departments and decision-makers, providing a foundation for risk management strategies that prioritize highly reliable predictions. We further present visualizations of FireCat’s predictions for Miami-Dade and Broward Counties (Figs. \ref{fig:step2_accuracy}g, h, i), alongside the observed values. The model demonstrates strong spatial fidelity in accurately classifying fire consequence categories.

\subsection{Event-level fire incident factors dominate the fire consequence risk levels.}
\label{sec:consequence_event}
The contributions of factors in the FireCat are evaluated using SHAP values \cite{shapley1953value,luan2023data}, where higher values indicate a stronger positive impact on the outcome. factors are grouped into two sets: incident-specific fire incident factors (derived from fire incident reports) and local factors (including social determinants, business proportions, building inventories, and weather conditions). For each set, the top eight factors influencing fire consequence levels are identified and visualized in Fig. \ref{fig:step2_shap}a (the complete variable list and corresponding description are shown in Supplementary Section 1). Across all targets, incident-specific factors (in light color shading) consistently exhibit relatively higher SHAP values than local factors, underscoring the dominant role of incident-specific characteristics in predicting fire consequences. Below, we organize key findings by dividing incident-specific fire incident factors into three categories: 1) fire factors, 2) building factors, and 3) spatial-temporal factors.

Fire factors emerge as some of the most influential predictors of fire consequences (Figs. \ref{fig:step2_shap}b, c, d). Specifically, the location of fire origin significantly influences all three fire consequences. Fires starting in 'storage areas', exhibit a heightened risk of spread, while those igniting in assembly areas like classrooms and arenas, though less prone to spreading, are associated with higher risks of injury. Fires starting in technical processing areas tend to elevate the economic loss risk. Furthermore, ignition-related factors played a crucial role in predicting fire consequences. While equipment failure generally shows a reduced severity of fire spread and injuries, human-caused ignitions, particularly those involving individuals with physical disabilities, significantly increase injury risk.
Finally, the heat source emerges as a key factor in predicting fire consequence risks. Fires involving multiple heat sources showed the highest contribution to the elevated fire consequence risks compared to other heat sources.


 The presence or absence of fire detectors and AES emerge as a critical factor in determining both fire spread and injury risk, with buildings lacking detectors or AES exhibiting heightened vulnerability (higher SHAP values as shown in Figs. \ref{fig:step2_shap}b, c and Supplementary Fig. S8). However, only 6\% of buildings in the fire reports are equipped with AES, resulting in a relatively low average SHAP value, which prevents it from ranking among the top predictors.
Furthermore, building construction materials also play an important role in shaping fire consequences. Fires that ignited from structural components are associated with higher fire spread risk.
These findings reinforce the importance of improving fire safety standards in high-risk building types and enforcing the adoption of fire detectors in both existing and new buildings to mitigate risks.

 \begin{figure}[htpb]
    \centering
    \includegraphics[width=0.9\linewidth]{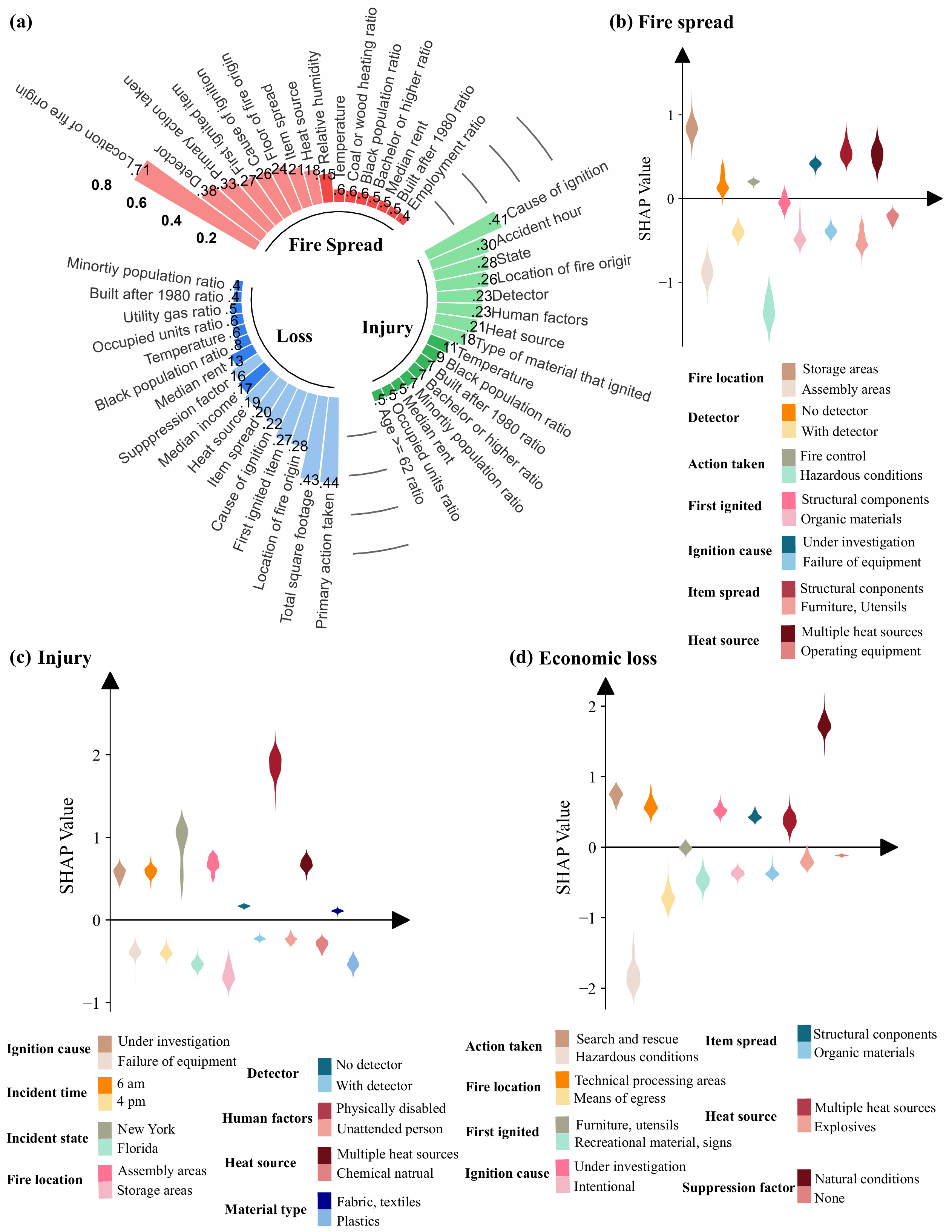}
    \caption{\small\textbf{Factor importance analysis of FireCat for predicting fire consequences based on SHAP values.} \textbf{(a)} Top eight most important incident-specific factors and local factors of the FireCat. The length of the bars represents the importance of factors, while the color shading distinguishes the incident-specific (light) and local (dark) factors. \textbf{(b, c, and d)} Sub-category of each incident-specific factor with the highest positive and negative influence on fire spread, human injury, and economic loss. Sub-categories with positive (negative) SHAP tend to increase (decrease) the predicted fire consequence levels. The complete spectrum of SHAP values for each sub-category within the incident-specific factors is detailed in Supplementary Figs. S5–S7.}
    \label{fig:step2_shap}
\end{figure}

Figs. \ref{fig:step2_shap}b and c reveal spatiotemporal variations in fire consequences, with injury risk peaking in the early morning (e.g., 6 a.m. fires exhibit higher risk than 4 p.m. fires, likely due to reduced alertness) and significant geographic differences in both human injury and economic loss. Notably, New York presents the highest risk of fire-related injuries, while Florida demonstrates the lowest. These findings underscore the need for tailored fire prevention strategies that prioritize high-risk communities and address specific regional and temporal vulnerabilities.

\begin{figure}[htpb]
    \centering
    \includegraphics[width=0.8\linewidth]{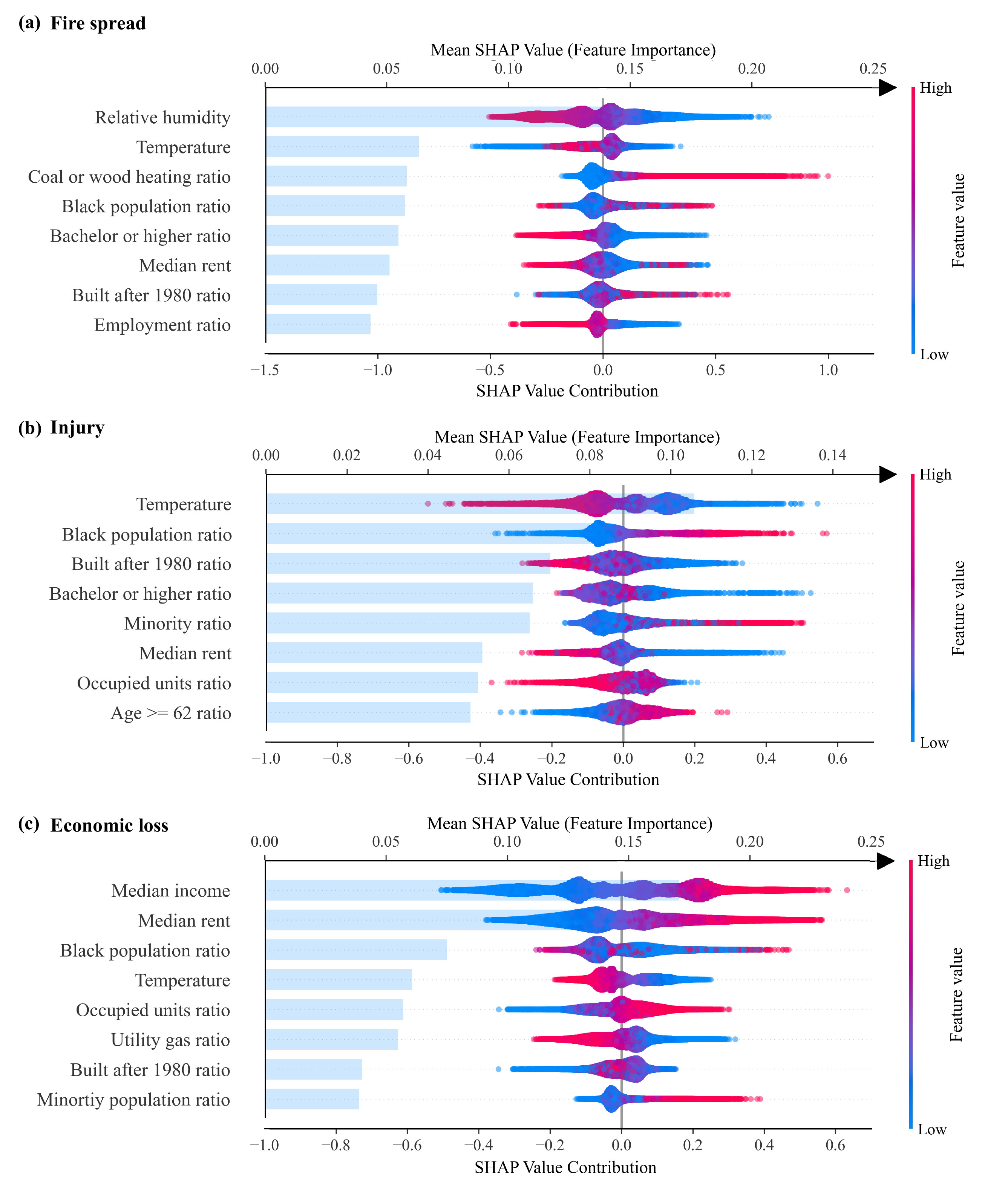}
    \caption{\small \textbf{SHAP value for top eight local factors influencing consequence levels of \textbf{(a)} fire spread, \textbf{(b)} human injury, and \textbf{(c)} economic loss.} 
The y-axis lists the names of the eight factors, ranked by their mean SHAP values (light blue bars, top x-axis), which represent each factor's overall importance in the model. The bottom x-axis displays SHAP value contributions, showing whether a factor increases (positive SHAP) or decreases (negative SHAP) the predicted fire consequence levels. A color gradient (blue to red) represents the factor's actual value: blue indicates low values, while red indicates high values, corresponding to the factor's data range. For example, in the context of fire spread, low relative humidity (blue) strongly increases the consequence level, while high humidity (red) reduces it. 
    }
    \label{fig:step2_shap_num}
\end{figure}

\subsection{Climate and socioeconomic local factors also influence the consequences of a fire}
Fig. \ref{fig:step2_shap_num} presents the SHAP values of the top eight local factors influencing the consequence levels of fire spread, human injury, and economic loss. The variables are ranked by their relative importance, as indicated by the light blue bars on the left, while the beeswarm plots provide detailed insights into the contribution of each factor across different conditions.

For fire spread (Fig. \ref{fig:step2_shap_num}a), relative humidity emerges as the most significant predictor, with lower humidity levels markedly increasing fire spread risk. For injury outcomes (Fig.\ref{fig:step2_shap_num}b), temperature and racial composition (Black population ratio) are the leading contributors, with higher temperatures and areas with a greater Black population ratio associated with increased injury risk. Educational attainment, represented by the Bachelor or higher ratio, is inversely associated with fire spread and injury risk. However, this relationship likely reflects broader socioeconomic and infrastructural factors at the community level rather than direct individual preparedness. Higher educational attainment is often correlated with greater annual income \cite{Braveman2011}, and potential is associated with greater investments in building safety measures, and enhanced proximity to emergency response services, all of which contribute to reducing fire risks. While individual awareness and preparedness may play a role, the observed association is more plausibly driven by community-level disparities in the built environment and public services.

Interestingly, while higher median rent is generally associated with reduced fire spread and injury risks (Fig. \ref{fig:step2_shap_num}a-b), it is positively correlated with economic loss from fire (Fig. \ref{fig:step2_shap_num}c), potentially reflecting greater property value in more affluent areas. Collectively, these insights underscore the complex interplay of local factors driving fire consequences, highlighting the importance of understanding community-specific vulnerabilities, even in the presence of dominant incident-level factors (as discussed in Section \ref{sec:consequence_event}).

\subsection{Partial dependence analysis of factor interactions}

\begin{figure}[htpb]
    \centering
    \includegraphics[width=0.85\linewidth]{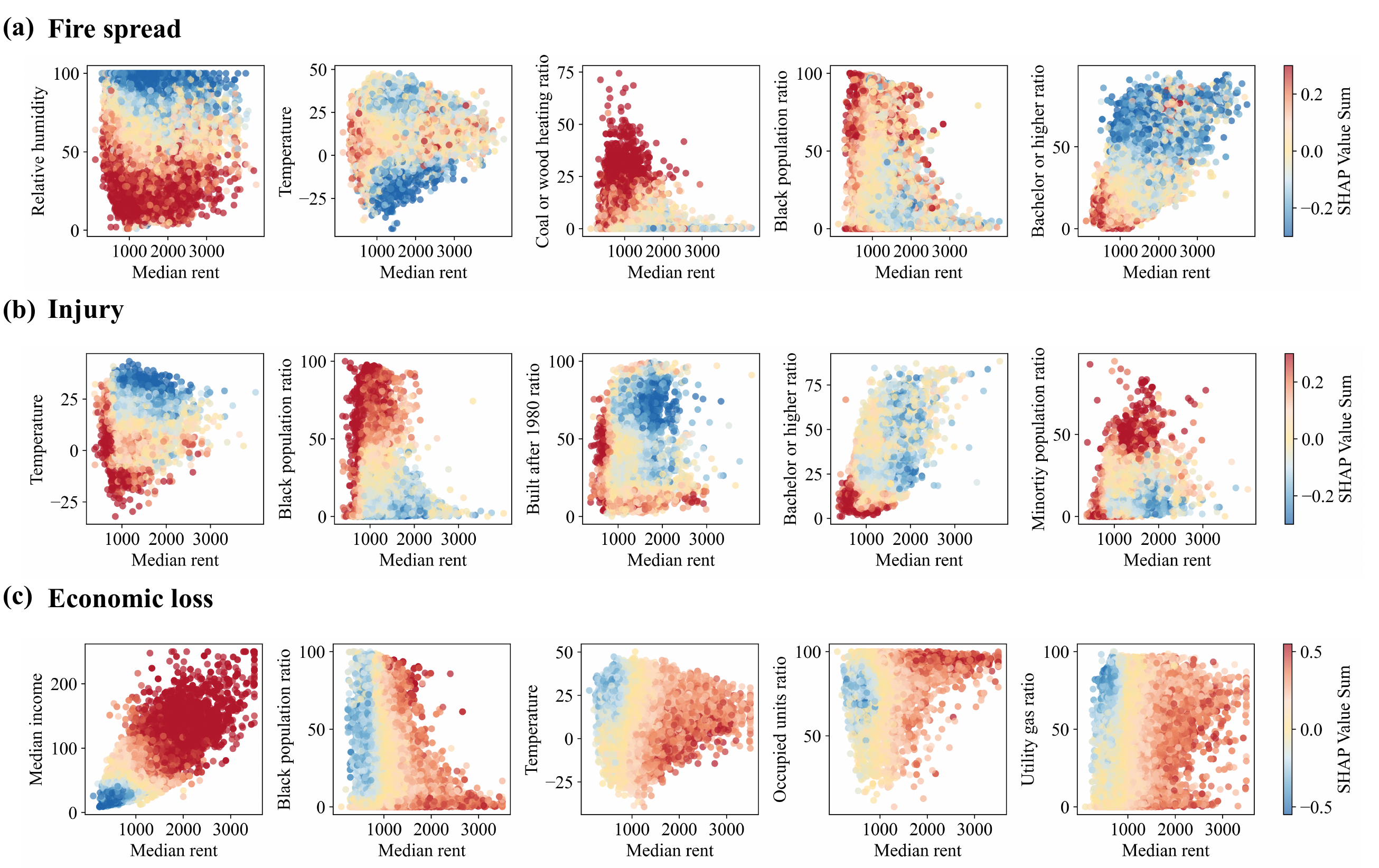}
    \caption{\small \textbf{Partial dependence between median rent and top 5 local factors in \textbf{(a)} fire spread, \textbf{(b)} injury, and \textbf{(c)} economic loss based on their SHAP value.} The horizontal axis represents the value of 'Median rent' in USD, whereas the vertical axis represents the value of other local factors. The color gradients in the figure indicate the partial dependence of the predicted target concerning a specific x-value and y-value combination.  Red color indicates a positive relationship with increased consequence level (higher SHAP values), while blue indicates a negative or lower contribution to the consequence level (lower SHAP values), with yellow representing a neutral or moderate effect. The complete spectrum of partial dependence across the top local factors is presented in Supplementary Figs. S9–S11.}
    \label{fig:step2_partial}
\end{figure}

To understand how median rent, which serves as a common proxy for socioeconomic conditions and consistently appears among the top eight predictors for all three fire consequences, interacts with other factors influencing fire consequence levels, we conduct a two-factor partial dependence analysis (Fig. \ref{fig:step2_partial}). This analysis reveals complex interactions between median rent and other key factors of fire consequence risks. For fire spread and injury risks, the influence of other predictors varies significantly based on median rent levels (Figs. \ref{fig:step2_partial}a, b).
In low-rent areas (below \$1,000 per month), these factors have a limited impact on variations in fire consequence risks. Conversely, in high-rent areas (above \$1,000 per month), vulnerable communities, characterized by building-related factors (e.g., fewer newer buildings, greater reliance on coal or wood heating) and social determinants (e.g., lower education levels, higher proportions of Black and minority populations), are associated with increased fire spread and higher injury rates.
Weather characteristics, such as relative humidity and temperature, also notably interact with median rent. While significantly impacting fire spread and injury consequence levels overall, their influence is negligible in lower-rent areas, suggesting that socioeconomic disadvantages may overshadow climatic effects in shaping fire consequences. Finally, Fig. \ref{fig:step2_partial}c indicates that local economic conditions primarily drive fire-related economic loss, specifically median income and median rent. This finding reinforces the strong link between property value and fire-related financial impact, as fires in higher-income areas tend to result in greater economic losses due to higher property valuations and asset exposure.

\section{Discussion}

Our framework integrates local and incident-specific data for enhanced fire risk analysis, complementing traditional building fire risk analysis frameworks that heavily depend on individual buildings and predefined scenarios and often overlook these critical sources of information. For instance, the effects of variation in local factors (e.g., education levels) and event-level factors (e.g., fire ignition location) on fire risks are not considered in traditional fire risk analysis frameworks or standards \cite{IBC,InvisibleUSFireProblem,NFPA557,EN1991-1-2}. To address this issue, our proposed model unlocks the full potential of nationwide building fire reports over the past decade for fire risk analysis, by matching the local information with each fire incident and adapting data-driven methods. As demonstrated in section \ref{firecat_performance}, the fire-consequence prediction model achieves a significant performance increase of at least 30.37\% in accuracy over the baseline, highlighting the value of incorporating additional local and incident-specific data for fire risk analysis. While our current model is a promising first step, it provides a critical stepping stone in fire risk analysis, which can be directly expanded to include an even broader spectrum of local and incident-specific factors, such as building design and human behavior data, further enhancing predictive accuracy and usefulness.

 Our building fire risk analysis framework provides critical insights into identifying fire-vulnerable communities shaped by local disparities and supports policymakers in developing tailored prevention strategies. By employing GAMs for fire occurrence analysis at the county-level (Fig. \ref{fig:step1_occ}) and SHAP value calculation for fire consequences analysis at the ZIP code level (Fig. \ref{fig:step2_shap_num}), the framework identifies how specific local factors contribute to the fire risk. 
For instance, we identify that high vulnerability in local social determinants and building inventories is associated with increased fire occurrence, fire spread, and injury consequence levels. We also notice a strong association between cooler and dryer weather with higher fire risks. These findings align with patterns documented in existing literature on a single aspect of fire occurrence or fire consequences with data at a specific region \cite{schmidt4757205machine,hossain2023analyzing,ku2024charactering}.
These factors point to vulnerabilities that may stem from systemic inequities, limited access to fire safety resources, outdated construction standards, poorly maintained buildings, and utilization of flammable building materials, affecting infrastructure and community preparedness. Such systemic inequities are also observed in other hazards \cite{ermagun2024high,hallegatte2016socioeconomic,anderson2023inequality}. 
These findings can help inform policy that ensures equitable access to fire safety resources and prioritizes modernizing outdated infrastructure, particularly in disadvantaged areas. Fire prevention efforts must also address local climate conditions and the unique local characteristics of each community to enhance fire resilience.

Our analysis reveals that the most influential incident-specific factors vary across different fire consequence targets (Fig. \ref{fig:step2_shap}a). Fire-related factors are the most dominant for fire spread and economic loss risks, while human-related factors have the most influence on human injury risk during a fire event. 
The difference stems from the fact that fire spread and economic loss are more dependent on the physical attributes of the fire and building, while human injury is more dependent on an individual's response. For example, the location of the fire origin can significantly affect fire growth and suppression, resulting in different fire spread and economic loss levels (Figs. \ref{fig:step2_shap}b, d). In contrast, people's awareness of fire and ability to evacuate are more influential for human injury (Fig. \ref{fig:step2_shap}c).
This differentiation highlights the multifaceted nature of fire incidents, emphasizing the need for tailored strategies in fire risk mitigation and response planning for diverse objectives, such as developing efficient evacuation strategies \cite{kinateder2015risk,sun2020bim}. 
Our models’ explainable factors and demonstrated predictive capabilities can provide decision-makers with a detailed understanding of fire risks and vulnerabilities within communities.


Our analysis highlights the significant role of fire detectors (Fig. \ref{fig:step2_shap}) and AES (Supplementary Fig. S8) in reducing fire consequence risks, especially for fire spread and injury. This finding supports the efficiency of regulations to mandate the installation and maintenance of advanced fire safety systems \cite{festag2023minimization,liu2024preparation,nivennesh2021autonomous}, particularly in high-risk or vulnerable communities, such as low-income neighborhoods, multi-family residential buildings, and facilities housing elderly or disabled individuals.
Policymakers can leverage these insights, within cost-benefit assessments \cite{van2023cost}, to update building codes and regulations, mandate retrofitting of existing buildings, and possibly implement subsidy programs for installing fire detectors and AES in disadvantaged areas, ensuring equitable access to fire safety resources. Additionally, public awareness campaigns could emphasize the life-saving benefits of these systems, encouraging voluntary adoption. 
By aligning policy with data-driven findings, decision-makers can better allocate resources, prioritize fire prevention strategies, and reduce disparities in fire safety outcomes. These targeted measures will not only mitigate fire risks but also promote safer, more resilient communities nationwide.

Standardized post-fire data collection is crucial for enhancing fire prevention and building resilience across states. While the current NFIRS dataset provides valuable insights into fire incidents, additional standardization and supplementation can further improve data quality \cite{barakey2009fire,anderson2018exploration}. Establishing consistent criteria for recording data will allow for more meaningful comparisons across jurisdictions, improve our understanding of fire events, and contribute to developing tailored fire prevention strategies. 
Additionally, incorporating elements of structural fire engineering into the data collection process can provide crucial information on how buildings respond to fires, which in turn helps inform future construction and retrofitting practices. Key data points could include structural damage status, the types of materials used in construction (e.g., combustible vs. non-combustible materials), and the age and type of building. These data would support the development of building codes that promote safer, more resilient structures, especially for high-risk building types.


This study has several limitations. First, our analysis identifies associations rather than causal relationships between fire risk factors and fire outcomes. The relationships identified may be influenced by unmeasured confounding factors or underlying structural conditions that shape fire risks. Future research should explore causal inference approaches, such as controlled experiments, to better understand the causal impact of these factors on fire occurrence and consequences. Second, to facilitate analysis, textual information from fire incident reports is converted into numerical inputs, potentially overlooking nuanced information embedded within the text. Future research can leverage large language models (LLMs) to capture the contextual and semantic richness of this textual data. Third, certain categorical factors with numerous subcategories (e.g., location of fire origin, property usage) are aggregated into broader groups based on the NFIRS codebook to ensure consistency and analytical tractability. While necessary, this may have reduced granularity and masked subtle patterns within the original data. Finally, the analysis assumes complete reporting of fire events within each ZIP code in the dataset. While necessary for analysis, this assumption may affect the representativeness of data for specific regions. Further research can explore methods to address potential reporting biases and incorporate more granular categorical information to enhance the model's accuracy and interpretive power.
\label{sec:discussion}

\newpage 
\def\XX{\mathbb{X}}
\section{Methods}
\subsection{Data}
\label{data}

NFIRS, the world’s largest national database of fire incident information, is used by participating US fire departments to uniformly report details on all types of fire to which they respond. At present over 24,000 fire departments report to the NFIRS system, which represents about 75 percent of all fires in the U.S.  For this analysis, we extract all the building fires from the year 2012 to 2022, reported from 2,976 unique counties (out of 3,144) and 29,244 unique ZIP-code (out of 41,642). The number of reported incidents is reduced from six million to over one million by restricting the cases to building fires in enclosed and normal-use buildings with fires extended beyond items that ignited (e.g., fires confined to cooking pots). From the NFIRS database, we obtain the incident-specific information of each fire event, including burned building data, fire data, fire fighting data, and spatial-temporal data. From the sptial-temporal data provided in each report, the ZIP code level location information and hour-level time information can be extracted. By matching the ZIP code and year of the incident with the U.S. Census Bureau \cite{USCensusBureau} database, specific demographic, socioeconomic, and building inventory data can be obtained. This linkage allows for extracting factors such as population density, income levels, education levels, and other community characteristics that may influence the vulnerability to fire risk. The time information enables the integration of historical weather data at the time and location of each incident. These data are then used as explainable factors to help understand the building fire risks.

\subsubsection{Event-level incident data}

\textbf{Building data}: Building characteristics can significantly affect the fire risk. The International Building Code (IBC) considers the construction material, building height, and area as key parameters that affect the requirements for fire safety measures\cite{IBC}. The European building code \cite{EN1991-1-2} also defines the severity of fires based on building usage. In the U.S. code, fire occurrence is also considered conditional to property usage \cite{NFPA557}. 


Our study focuses on building fires, including normal-used enclosed buildings, while excluding mobile structures, open structures, underground structures etc, to better understand fire risks in common residential, commercial, or industrial buildings. Factors include information about the physical characteristics of the building involved in fire incidents, such as property usage (e.g., office, multifamily dwelling, warehouse, etc), the number of stories below and above ground, the number of units, the total square footage, and whether the building is equipped with a fire detector or AES. 

\textbf{Fire data}: The fire's growth and spread affect its consequences.  Different materials exhibit varying heat release rates (HRR). Materials with a high heat release rate generate more energy per unit time, leading to a rapid increase in gas temperature within the fire compartment. This elevated gas temperature not only intensifies the fire's growth but also accelerates the ignition of adjacent materials, contributing to a faster spread of the fire. Therefore, the location of fire origin and material contribution to fire ignition can be very informative in evaluating the fire risk. 


The considered fire data includes the cause of fire ignition (e.g., intentional, failure of equipment, act of nature, etc), the location of fire origin (e.g., bedroom, kitchen, living room, etc), the first ignited item, the material type of the first ignited item, items and materials that contribute most to fire spread, the heat source of the fire, the fire growth timeline (e.g., first floor, second floor, etc), factors contributing to the ignition (e.g., misuse of material or product, heat source
proximity to combustibles, etc), and human factors contributing to the fire ignition (asleep, possibly impaired by alcohol or drugs, etc). 

\textbf{Fire fighting data}: Numerous research findings have highlighted the influence of firefighting operations, with factors such as fire department response time, on fire consequences  \cite{challands2010relationships}. Rapid fire department response time can prevent the transition from a localized fire to a fully developed fire (also known as flash-over), which significantly increases the fire's intensity and spread \cite{bishop1998fires}.   

%
To better understand how the fire fighting process affects the building fire risk, three factors are considered, including the primary action taken by the firefighters, the fire department response time (from alarm time to arrival time), and the factors that contributed to the growth, spread or suppression of the fire (e.g., trouble finding locations).

\subsubsection{Social determinant data}

\textbf{Demographic data}: The demographics data can help identify the vulnerability of communities against various hazards, such as earthquakes, hurricanes, and fires \cite{ermagun2024high, hallegatte2016socioeconomic}. Incorporating demographic data such as race- and age-related information and the percentage of the population with a disability can help understand how they influence fire risk and identify communities more susceptible to fire incidents.

\textbf{Socioeconomic data}:
Socioeconomic indicators like education level, employment ratio, health insurance coverage ratio, median income, and rent level are important for assessing the economic status of communities, which can influence the community's resilience towards fire risks \cite{hallegatte2016socioeconomic}. Our study considers these socioeconomic factors to understand the association between the community's socioeconomic factors and its fire risks. 

\subsubsection{Business proportion data}
Besides the general socioeconomic data, we consider the business data, such as the proportion of establishments in transportation and storage (with NAICS codes 48-49 and 81) and industrial (with NAICS codes 22, 23, and 31-33) in each county, to understand how these factors affect the fire occurrence risks.

\subsubsection{Building inventory data}
As building code requirements on fire safety measures evolve over time, building characteristics such as the date of construction are relevant. Considering that a large portion of building fires occur in residential properties \cite{USFA}, the heating source used within these buildings is also an important variable. 
Additionally, unoccupied or vacant buildings are vulnerable to fire hazards according to \cite{schachterle2012proximity}. We therefore consider building inventory data, including the proportions of buildings constructed after 1980, the proportion of buildings with certain heat sources (gas, wood, etc.), and the building occupied ratio, to in the fire risk analysis.

\subsubsection{Weather condition data}
Weather conditions at the time of fire incidents can affect the development of fires \cite{jain2022observed}. We incorporate climate data, including temperature, relative humidity, wind speed, and Palmerr Z-index, into fire risk analysis to obtain a comprehensive understanding of weather conditions on fire risks, support accurate predictive models, and aid in informing weather-specific fire prevention and response strategies.

\subsection{Fire Occurrences}
\label{sec:occurrences}
We examined the spatial and temporal variation of fire occurrences, using the county-level annual fire incidence rate per 100,000 building units as the target variable. The fire occurrence analysis is designed to move beyond individual building-level or fire incidence-level characteristics and provide a broader understanding of community-level fire occurrence risk.

\subsubsection{Variable Selection}
\label{method:step1_variables}
In this study, we utilize county-level spatial-temporal data as input for our model. We select two demographic variables: the Black population ratio, which reflects the racial composition of the county, and the senior population ratio, which reflects the age distribution. For socioeconomic factors, we use the percentage of the population with a bachelor’s degree or higher and the urban population ratio to measure the community’s socioeconomic status.  To capture characteristics of the building stock, we include two building stock variables: the occupied unit ratio, indicating the state of building maintenance, and the proportion of buildings constructed after 1980, representing the age of the building infrastructure. Additionally, we incorporate two variables related to the economic activity in the county: the ratios of establishments in the transportation and storage sector and the industrial sector. Finally, we include two climate variables: the average monthly temperature and the Palmer Z-Index, which capture the key environmental factors. 

\subsubsection{Study Design}

Generalized Additive Models (GAMs) are employed to model the annual county-level fire incidence rate. The motivation for using GAMs is rooted in their flexibility in capturing nonlinear relationships between the predictors and the target variable, which is crucial given the complex interplay of factors influencing fire occurrences. GAMs extend linear regression by allowing for smooth, non-parametric functions of the predictors, providing a way to model more intricate patterns without imposing a strict functional form.  By fitting smooth functions to each predictor, GAMs provide an interpretable approach that can reveal important insights into the relationships between predictors and fire incidence rates.

Our proposed GAM is formulated as follows:
\begin{equation}\label{eq:gam1}
    \begin{aligned}
    log(E(I_c^m))  =& \ \beta_0 +  \beta_s \cdot \text{factor(state)} + \sum_{i = 1}^nf_i(X_{i,c}),
    \end{aligned}
\end{equation}
where $I_c^m$ represents the fire incidence rate for county $c$ and month $m$, $\beta_0$ is the intercept, $\beta_s$ are the parameters for the state-specific ($state$) fixed effects, the non-parametric function $f_i$ are the smooth function for other county-level variables ($X_{i,c}$) including ten variables depicted in section \ref{method:step1_variables}. Detailed formulations are presented in Supplementary Information Section 2.

The fire incidence is fitted with a Gamma distribution, which assumes:
\begin{equation}\label{eq:gam2}
    \begin{aligned}
    I_c^m \sim \Gamma(\mu, \theta),
    \end{aligned}
\end{equation}
where $\mu = E(I_c^m)$ is the mean fire incidence rate, and $\theta$ is the dispersion parameter. The log-link function ($g(\cdot)$) ensures that the predicted mean of $I_c^m$ remains positive:
\begin{equation}\label{eq:gam3}
    \begin{aligned}
    g(\mu) = log(\mu).
    \end{aligned}
\end{equation}

To comprehensively analyze the factors influencing fire incidence, we employ three distinct sets of GAMs. While each set utilizes the same predictor variables, they are applied to different subsets of the data to capture variations in fire occurrence risk across spatial and temporal scales. The full list of variables used in this model is presented in Supplementary Section 1.1.

\textbf{National GAM:} A single GAM is fitted to the entire national dataset. This model serves as a baseline, identifying the general relationships between the predictors and fire incidence across the United States.

\textbf{Seasonal GAMs:} Four separate GAMs are constructed, one for each season (spring, summer, autumn, winter). This approach allows for the identification of seasonal trends and variations in the influence of fire occurrence.

\textbf{Regional GAMs:} Four regional GAMs are developed, representing the Northeast, Midwest, South, and West regions of the United States. The states included in each region are detailed in Supplementary Section 1.2. This allows for the exploration of spatial heterogeneity in the relationships between predictors and fire occurrence.

\subsection{Fire Consequences}
\label{sec:consequences}

In this section, we describe our methodology for modeling three aspects of fire-related consequences: 1) fire spread, 2) fire-induced human injury, and 3) fire-induced economic losses. We fit a CatBoost model to predict each of these risk categories, leveraging a range of social determinants, building inventories, and fire incident-related predictors. The full list of variables used in this model is presented in Supplementary Section 1.3. The model's performance is evaluated using accuracy, F-1 score, precision, mean squared error (MSE), weighted mean squared error (WMSE), Brier score, and rank probability score (RPS) for each risk category, and we further assess the impact of each predictor using SHAP (Shapley Additive exPlanations) values \cite{lundberg2017unified} to understand the contribution of individual factors towards each risk.

\subsubsection{Targets Design}
\label{method:step2_targets}
\textbf{Fire spread status:} The fire spread risk is defined based on categorical outcomes related to fire propagation, which is provided in each NFIRS report. This target evaluates the extent to which a fire can propagate within a building or surrounding environment. This metric is essential for understanding the potential escalation of a fire incident, as uncontrolled fire spread can result in greater damage, impact neighboring areas, and complicate firefighting efforts.

\textbf{Human injury:} The fire-induced injury risk is defined based on the injury severity and the number of injuries that are provided in the NFIRS report for each fire incident. To effectively account for both factors, each injury severity level is assigned a specific coefficient: 0.003 for minor injuries, 0.047 for moderate injuries, 0.266 for severe injuries, 0.593 for critical injuries, and 1 for fatalities. These coefficients are adjusted based on literature \cite{zhuang2017total}. By combining these coefficients with the total number of injuries per fire event, we calculate a weighted injury value that captures the overall impact of the incident, balancing the severity of injuries with their frequency. The detailed equation for calculating the combined injury indicator is presented in Supplementary Section 1.3. This approach enables a more nuanced categorization of fire incidents based on injury impact. Events are classified into risk levels by quantile: the bottom 40\% quantile represents low injury risk, the 40-75\% range as moderate, and values above the 75\% quantile as high risk. Understanding fire-induced injury risks is vital for assessing the human impact of fire incidents and designing targeted interventions to improve safety measures and reduce injury rates in vulnerable populations.

\textbf{Economic loss:} The fire-induced economic loss risk is defined based on the recorded property and content loss in the NFIRS report for each fire incident. Loss values are converted to 2022 value according to the CPI \cite{cpi2024} and segmented into three groups according to their quantiles: events falling within the bottom 40\% quantile are considered low economic loss, while those between the 40-75\% range are classified as moderate. Events with economic losses above the 75\% quantile are designated as high risk. Understanding fire-induced economic loss is essential for estimating the economic consequences of fire incidents, identifying high risk areas for resource allocation, and informing insurance and mitigation strategies.

\subsubsection{Study Design}
\label{step2_study_design}
\textbf{Model:} To model fire-related risks, we develop FireCat, a predictive model built on CatBoost \cite{prokhorenkova2018catboost}, a gradient boosquantilesting decision tree algorithm optimized for datasets with numerous categorical factors. CatBoost is chosen due to its ability to handle categorical data natively, making it well-suited for our dataset, which includes various categorical fire incident-related factors (e.g., cause of fire ignition, the area of fire ignition, and property usage). Unlike traditional gradient boosting frameworks that require one-hot encoding, CatBoost uses an efficient combination of ordered boosting and target encoding, which reduces computational overhead and minimizes the risk of overfitting. Additionally, as a tree-based model, CatBoost effectively manages correlated factors. FireCat is designed to predict fire consequences from three perspectives: spread risk, property loss, and human injury.

\textbf{Evaluation metrics:} Three CatBoost prediction models are constructed using fire incident data, spatio-temporal data at the ZIP code level, and climate data at the time of the fire events (including 50 explanatory factors) and are fitted to the three targets defined in Section \ref{method:step2_targets}. The data are split into two sets: 1) Training and validation set (70\%): This portion of the data is used to train the CatBoost models and optimize their parameters for optimal performance. 2) Testing Set (30\%): This portion of the data is held back to evaluate the final model's performance on unseen data. Model performance is assessed using seven key metrics: 1) Accuracy: Measures the overall correctness of the model's predictions. 2) Precision: The proportion of correctly predicted positive instances among all predicted positives, reflecting accuracy in identifying true positives. 3) F1-score: The harmonic mean of precision and recall, balancing false positives and false negatives. Higher values indicate better classification performance. 4) Mean Squared Error (MSE): Measures the average squared difference between predicted and actual values. Lower values indicate more accurate predictions. 5) Weighted Mean Squared Error (WMSE): A weighted version of MSE that accounts for varying importance of errors. Lower values indicate better performance in weighted scenarios. 6) Brier score (BS): Evaluate the accuracy of probabilistic predictions by calculating the mean squared error between the predicted probabilities and the actual outcomes. A lower Brier score indicates better predictive performance. 7) Ranked Probability Score (RPS): Assesses the accuracy of probabilistic predictions for ordinal outcomes. It measures the cumulative squared error between the predicted and observed cumulative probabilities. Like the Brier score, a lower RPS indicates better performance.

The formulation for BS follows:
\begin{equation}
    \text{BS} = \frac{1}{N} \sum_{i=1}^{N} \sum_{k=1}^{K} (P(\hat{y}_i=k) - o_k^{(i)})^2
    \label{eq:Brier}
\end{equation}
where $P(\hat{y}_i=k)$ is the forecast probability for the $i$-th data point in the $k$-th class, and $o_k^{(i)}$ is the one-hot encoded ground truth, with a value of 1 for the targeted class and 0 for all other classes. 

The formulation for RPS is as follows:
\begin{equation}
    \text{RPS} = \frac{1}{N-1} \sum_{i=1}^{N} \sum_{k=1}^{K} (P(\hat{y}_i\leq k) - \bar{q}_k^{(i)})^2 ,
    \label{eq:rps}
\end{equation}
where $P(\hat{y}_i\leq k)$ and $\hat{q}_k^{(i)}$ are the predicted and observed cumulative probability distributions for the $i$-th data point in the $k$-th class, and $K$ is the number of categories in the ordinal outcome.

By employing these comprehensive evaluation metrics, we gain a robust understanding of the models' predictive capabilities, including both classification accuracy (accuracy and confusion matrix) and the reliability of their probabilistic forecasts (BS and RPS).

\textbf{Baseline model:} To further evaluate the performance of our FireCat model for fire risk prediction, we implement a baseline probabilistic model that relies solely on the observed target distribution in the training data. Specifically, the baseline model assumes no relationship between input factors and the target variable and instead uses the empirical distribution of the target values from the training set as its prediction.

For all instances in the testing set, the baseline predicts the same target distribution observed during training. This approach represents the most reasonable guess in the absence of a predictive model, providing a natural reference point to assess the added value of FireCat in capturing meaningful feature-target relationships for fire risk prediction.

\textbf{Factor attribution:} The impact of each factor is evaluated using SHAP values, which provide a game-theoretic approach to explain the output of the model \cite{shapley1953value}. SHAP values quantify the contribution of each factor to the prediction by attributing the difference between the actual prediction and the mean prediction to each factor. For categorical factors, SHAP values help identify which specific category has the most significant impact on the prediction, while for continuous factors, they measure how changes in the factor value influence the prediction.

\subsection{Software}
The analysis in Section \ref{sec:occurrences} was conducted using R (version 4.4.1) with packages mgcv (version 1.9.1), gratia (version 0.9.2), and tidy (version 1.3.1). The analysis in Section \ref{sec:consequences} was conducted using Python (version 3.10.12) with packages catBoost (version 1.5.2)and numpy (version 1.26.4).

\newpage 

\section{Data availability}
 All data utilized in this study derive from publicly accessible sources. Fire incident data were collected
 from the National Fire Incident Reporting System (NFIRS) \cite{NFIRScitation} in the U.S. Fire Administration. The county level and ZIP code level social determinant data, business proportion data, and building inventory data were sourced from the U.S. Census Bureau \cite{USCensusBureau}. The monthly weather data including the average temperature and palmer z-index were from the National
Oceanic and Atmospheric Administration (NOAA) \cite{noaa_climate_data}. The hourly weather data including the temperature, humidity, wind speed, and precipitation were collected from the Meteostat database \cite{meteostat}. Source Data for Figures 1, 2, 3, 4, and 5 are available with this manuscript.
\section{Code availability}
All codes are written using Python 3.10.12. Codes are publicly accessible at an archived repository and \url{https://github.com/Chenzhi-Ma/Fire_risk}.
\bibliography{reference}
\section{Acknowledgment}
This research received no specific grant from any funding agency in the public, commercial, or not-for-profit sectors.
\section{Author Contributions}
C.M., H.D., S.L., and T.G. conceptualized and designed the study. C.M. and E.D. collected data. C.M. and H.D. processed
 the data and designed model. C.M., H.D., S.L., and E.D. prepared the figures. C.M., H.D., and S.L. analyzed the results. C.M., H.D., and S.L. wrote the initial draft. L.M.G. and T.G. provided guidance and feedback for the
study. L.M.G. and T.G. revised the manuscript. All authors prepared the final version of the manuscript

\section{Competing interests}
The authors declare no competing interests.


\end{document}